# Airway segmentation from 3D chest CT volumes based on volume of interest using gradient vector flow

Qier Meng [#1], Takayuki Kitasaka [*2], Masahiro Oda [#1], Junji Ueno[4], Kensaku Mori[4,1]

[#1] *Graduate School of Information Science, Nagoya University*
qmeng@mori.m.is.nagoya-u.ac.jp
moda@mori.m.is.nagoya-u.ac.jp

[*2] *School of Information Science, Aichi Institute of Technology*
kitasaka@aitech.ac.jp

[3]*Department of Diagnostic Radiology, Graduate School of Health Sciences,*

*Tokushima University*
ueno@medsci.tokushima-u.ac.jp

[4]*Strategy Office, Information & Communications,*

*Nagoya University*
kensaku@is.nagoya-u.ac.jp

*Abstract—* Some lung diseases are related to bronchial airway structures and morphology. Although airway segmentation from chest CT volumes is an important task in the computer-aided diagnosis and surgery assistance systems for the chest, complete 3-D airway structure segmentation is a quite challenging task due to its complex tree-like structure. In this paper, we propose a new airway segmentation method from 3D chest CT volumes based on volume of interests (VOI) using gradient vector flow (GVF). This method segments the bronchial regions by applying the cavity enhancement filter (CEF) to trace the bronchial tree structure from the trachea. It uses the CEF in the VOI to segment each branch. And a tube-likeness function based on GVF and the GVF magnitude map in each VOI are utilized to assist predicting the positions and directions of child branches. By calculating the tube-likeness function based on GVF and the GVF magnitude map, the airway-like candidate structures are identified and their centrelines are extracted. Based on the extracted centrelines, we can detect the branch points of the bifurcations and directions of the airway branches in the next level. At the same time, a leakage detection is performed to avoid the leakage by analysing the pixel information and the shape information of airway candidate regions extracted in the VOI. Finally, we unify all of the extracted bronchial regions to form an integrated airway tree. Preliminary experiments using four cases of chest CT volumes demonstrated that the proposed method can extract more bronchial branches in comparison with other methods.

## I. Introduction

The lung cancer and chronic pulmonary disease are currently the leading causes of mortality worldwide. Anatomically, human airways appear as a tree-like branching structures of tubes that enable the air flow into the lungs through the trachea. An airway tree usually consists of more than 10 generations of branches. Due to the complex tree-like branching structure of airway tree, the inspection of disease often becomes difficult. However, obtaining a complete 3-D airway tree from a CT volumes is quite important and challenging. Several studies have focused on segmenting the airway tree from a chest CT volume. The most commonly used method is 3D region growing[1]. Such so-called "region-growing" methods, however, can not extract the peripheral branches due to the partial volume effect (PVE) and often lead to leakage. Lo et al[2] proposed a machine learning based method by extract the features based on intensity information and anatominal knowledge informations. This method can avoid the leakage effectively, but the periphral bronchi can not be extracted effectively. In order to track the smaller bronchi from a CT volumes, the tubular enhancement filters are utilized[3]. Although these methods can increase the detection rate well, the false positive rate increases simultaneously.

In this paper, we propose a new airway segmentation method from 3D chest CT volumes based on iterative extension of VOI[4] enhanced by CEF using GVF. To avoid the false positive (FP) occured in each VOI, a leakage detection function is utilized. We aim to improve the accuracy of the airway segmentation by tracing the airway region from the trachea and decrease the FP regions meanwhile.

## II. Method

### A. Overview

Figure 1 shows a flowchart of the proposed method. The proposed method consists of eight steps: 1) trachea extraction, (2) VOI setting, (3) pre-processing, (4) CEF application to VOI image, (5) bronchus region detection and leakage removement, (6) bifurcation detection and child VOI placement, (7) reconstruction of airway tree.

### B. Trachea extraction

Initially, the trachea region is obtained from a 3D chest CT volume by using a region growing based method[1] starting at a given seed point.

### C. VOI setting

In order to trace the airway, a VOI is defined as a local processing area along the direction of the bronchial branch running direction. The size of VOI is defined according to the bronchial branch radius. The detailed information is in Ref[5].

### D. Bronchus region detection and leakage removement

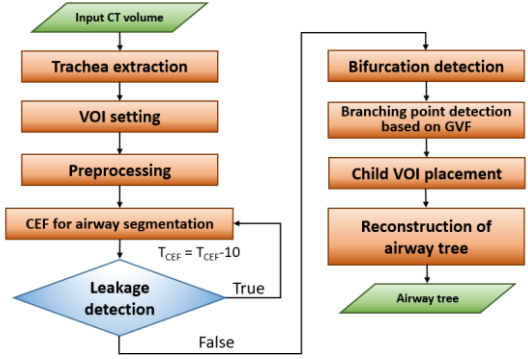

Fig.1. The flowchart of the proposed method

When a new VOI image is generated, a sharpening filter based on the Laplacian of Gaussian is utilized to enhance the VOI image to reduce the effect of PVE. After that, in order to extract the bronchial region from the VOI, the CEF is applied to the VOI image. The detailed definition of CEF can be seen in Ref[6]. After the candidate bronchial regions detected by CEF, there are some false positive (FP) regions in the results. In order to remove the FP regions, two steps are utilized: firstly, we compute the number of connected components by the labelling process. Then a leakage detection function is used to remove the leakage in the VOI by measuring the ratio $S_{ratio}$ between the bronchial area and VOI surface area. Then we extract the contour of a bronchial branch on the fronted surface of VOI extension. The circularity of the contour will be used to estimate the shape[7]. This circularity is also used as the measurement in the leakage detection function.

### E. Furcation detection

The furcation regions are detected by analysing the number of the connected components appearing on the surface of the VOI image. We classify the extracted voxels on the surface by 3D labelling process, and the number of voxels of each label is measured.

### F. Branching point detection and child VOI placement

If a bifurcation or trifurcation is detected, the child VOIs are needed to be generated to further trace the airway branches. And the branching point is needed to be defined to place the new child VOIs. In the proposed method, the GVF[8] is utilized for defining the branching point. By calculating the GVF, the initial vector field $F = -\nabla(G_\sigma * I)$ is computed and normalized as:

$$F^n(x) = \frac{F(x)}{|F(x)|} \frac{\min(|F(x)|, F_{max})}{F_{max}} \quad (1)$$

for every voxel, where I is the original CT volumes and $G_\sigma$ is a Gaussian filter kernel at scale $\sigma$. After obtaining the intial vector field, the GVF is calculated which is defined as:

$$E(V) = \iiint_\Omega \mu |\nabla V(x)|^2 + |F^n(x)|^2 |V(x) - F(x)|^2 dx \quad (2)$$

Figure 2 shows the illustration of GVF using a 2D cross section of an airway branch and the corresponding magnitude map.

In order to extract the airway centerline and identify the branching point of each bifurcation, a tubular-likeness function is given:

$$T(\mathbf{x}, r) = \frac{1}{2r\pi} \int_{\alpha=0}^{2\pi} \langle V(\alpha, r), D(\alpha) \rangle d\alpha \quad (3)$$

it is computed as the mean flow through the circle and depends on the radius $r$, where $V(\alpha, r)$ represents the GVFs vector at the circle point and $D(\alpha)$ defines a normal vector on the circle pointing towards its center. For obtaining the tubular-likeness, a circle is fitted to the data in the tubes cross-section plane. During the fitting procedure, the radius is steadily increased until the circle touches the edge of the object. The integral is approximately calculated by summing up over 32 discrete points on the circle. Figure 3 shows the illustration of the tubular-likeness procedure, and figure 4 shows the tubular-likeness result.

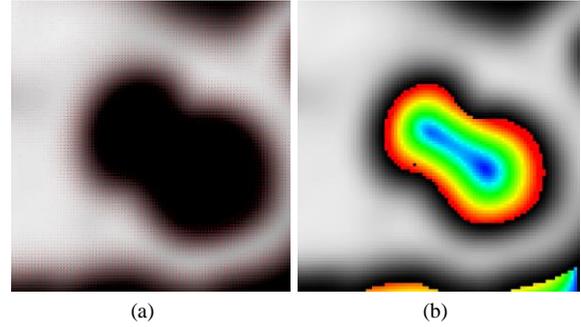

(a)          (b)

Fig.2. Illustration of the GVF using a 2D cross section of a airway structure. (a) Result of GVF field V(x), and the arrows indicate the vector direction. (b) Result of GVF magnitude map, and the red color indicates the higher value of the magnitude, and bule color indicates the lower value of the magnitude.

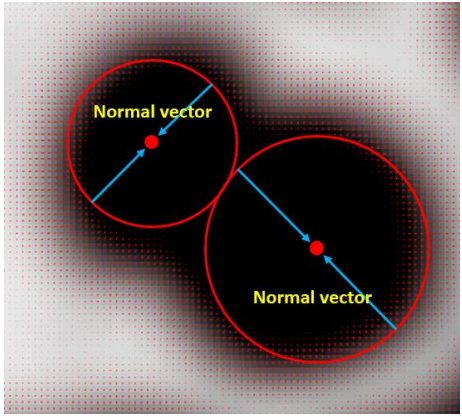

Fig. 3. The illustration of the tubular-likeness based on GVF.

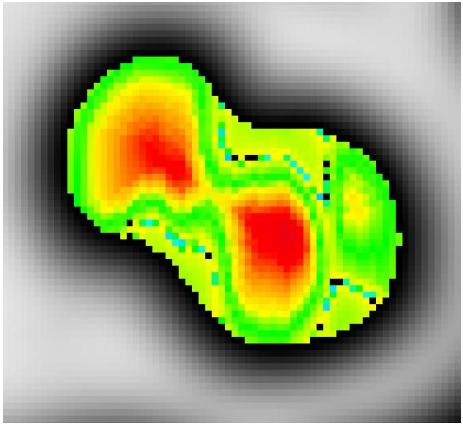

Fig. 4. The result of the tubular-likeness based on GVF. The red areas show the higher value of tubular-likeness, and the green areas show the lower value of tubular-likeness.

According to the Fig.2 (b) and Fig.4, it can be seen that the canter area can be enhanced. In Fig.2 (b) the center areas always show the GVF magnitude with low value, and in Fig.4 the centre areas always show the tubular-likeness with high value. In order to extract the centrelines, two given threshold values $t_m$ and $t_l$ are performed on the GVF magnitude and tubular-likeness for extracting the centreline candidate area. The regions where the GVF magnitude map is less than $t_m$ and more than $t_l$ are selected as the centreline candidate region. After the centreline extraction, a thinning process is performed on the extracted result. Figure 5 shows an example of final result.

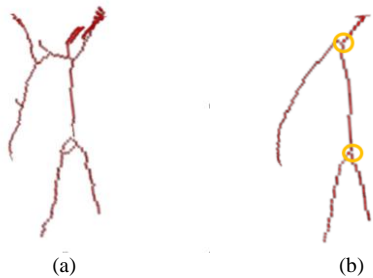

(a)            (b)

Fig. 5. An example of the centreline extraction result in VOI. (a) shows the thinning result of CEF in VOI by thinning approach. (b) shows the centreline result extracted by the proposed method.

After obtaining the centreline, a branching point identification method is used to obtain the branching point. The branching point is used to place the child VOI. The detailed information can be seen in Ref[5].

*G. Reconstruction of airway tree*

After finishing all processes, we construct the airway tree by projecting bronchial regions inside all VOI images into the original CT volumes.

### III. RESULT AND DISCUSSION

We used four cases of 3D chest CT volumes for evaluation of the proposed method. The parameter setting is as follows: the $β$ used in pre-processing is give as 0.05, the initial threshold value used in CEF is set as -800, and the $S_{ratio}$ is given as 0.33. The threshold value used in the tubular-likeness and GVF magnitude map are 500 and 50. The three indices were used for evaluation: (a) number of branches extracted, (b) ratio of extracted branches and total number of branches, and (c) false positive rate (FPR): number of extracted voxels which were not bronchial voxels actually. Table 1 shows the results of these cases. The ground truth data are created by manually.

TABLE I
RESULTS FOR THE PROPOSED METHOD FOR SEGMENTING THE AIRWAY TREE. THE FIRST LINE SHOWS THE RESULTS OF PROPOSED METHOD, AND SECOND LINE SHOWS THE PREVIOUS METHOD[4]

| Method | Average Extracted Numbers | Ratio of Extracted Number(%) | FPR(%) |
|---|---|---|---|
| Previous[4] | 95.25 | 76.5 | 0.76 |
| Proposed | 98.72 | 79.3 | 0.26 |

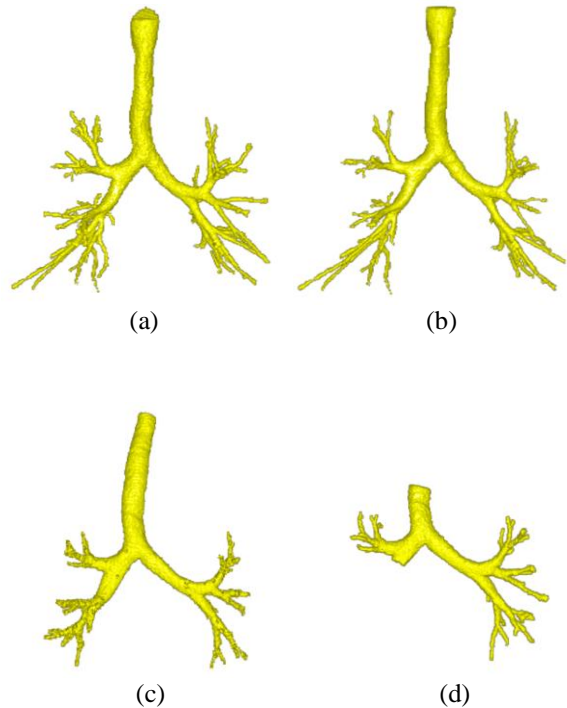

(a)            (b)

(c)            (d)

Fig.6. Two examples of the comparison results. The left column are the results of the proposed method. And the right column are the results of the previous method.

From the results shown in Table 1, the ratios of the extracted numbers of the bronchi can be increased around 2.8 percent, and the false positive rate can be decreased about 0.5 percent. Figure 6 shows two examples of the comparison results. The left column is the results of the proposed method, and the right column is the results of the previous method[2]. It clearly shows that the proposed method can extract more bronchi than the previous one.

Compare to the previous method, one advantage of the proposed method is that the presented centreline extraction can be easily implemented to obtain the skeleton of airway regions in the VOI which have the bifurcation. This can be achieved by extracting the centreline area in each VOI based on the GVFs medialness property. An example of the skeleton result has been shown in Fig.5 (b). For comparison, a skeleton with skeletonization approach on the CEF result directly is shown in Fig.5 (a). As can be seen, the skeleton extracted with the proposed method has higher quality, and the branching point can be easily identified, which give the position for placing the child VOI in the next level.

## IV. CONCLUSION

In this paper, we propose a new airway segmentation method from 3D chest CT volumes based on VOI using GVF. According to the results, the proposed method can perform segmentation results with high accuracy. And FP regions can be decreased. However, there remains possibility of generating false branches due to FP in the periphery of lung. Furthermore, there are still some peripheral branches missing due to the partial volume effect (PVE). Further investigation is required. In the future work, the GVF and tubular-likeness will be used for instructing the direction of the bronchi.

ACKNOWLEDGMENT

Parts of this research were supported by the MEXT, the JSPS KAKENHI Grant Numbers 25242047, 26108006, 26560255, and the Kayamori Foundation of Informational Science Advancement.

REFERENCES

[1] K.Mori and J. Hasegawa and J. Toriwaki and H. Anno and K. Katada, "Automated Extraction and Visualization of Bronchus from 3D CT Images of Lung," CVR Med'95, vol.1, pp. 542-548, 1995.
[2] Pechin Lo and Marleen de Bruijne., "Voxel classification based airway tree segmentation." Proceeding of SPIE on Medical Imaging 6914. 2008, 69141K.
[3] H. Yano and M. Feuerstein and T. Kitasaka and Jesper Johannes and K. Mori., "A novel ultrathin elevated channel low-temperature poly-Si TFT," *IEEE Electron Device Lett.*, vol. 20, pp. 569–571, Nov. 1999.
[4] T. Kitasaka and K. Mori and J. Hasegawa and J. Toriwaki, "A Method for Extraction of Bronchus Regions from 3D Branch Tracing and Image Sharpening for Airway Tree Chest X-ray Images by Analyzing Structural Features of the Bronchus," in *Forma*, 2002, vol.17, pp. 321-328.
[5] M. Feuerstein and T. Kitasaka and K. Mori., , "Adaptive Branch Tracing and Image Sharpening for Airway Tree Extraction in 3-D Chest CT," Second International Workshop on Pulmonary Image Analysis, Sep. 16, 2009.
[6] Hirano Yasushi and Rui Xu and Rie Tachibana, and Shoji Kido. "A Method for Extracting Airway Tree by Using a Cavity Enhancement Filter." Fourth International Workshop on Pulmonary Image Analysis, 91-99,2011.
[7] Maysa M. G. Macedo and Choukri Mekkaoui and Marcel P. Jackowski.. "Vessel Centerline Tracking in CTA and MRA Images Using Hough Transform," Progress in Pattern Recognition, Image Analysis, Computer Vision, and Applications, vol. 6419, pp. 295-302, 2010.
[8] Christian Bauer, Horst Bischof, and Reinhard Beichel, "Segmentation of Airways Based on Gradient Vector Flow" Second International Workshop on Pulmonary Image Analysis, Sep. 16, 2009.